\documentclass[conference]{IEEEtran}

\usepackage{amsmath,amssymb, graphicx}
\usepackage[ruled,linesnumbered]{algorithm2e}
\usepackage{algorithmic}
\usepackage{paralist} % also compact lists
\usepackage{color}
\usepackage{multirow}
\usepackage{rotating}
\usepackage{hyperref}
\usepackage[mathscr]{eucal} %% For \mathscr
\usepackage{amsbsy} %% For \boldsymbol
\usepackage{subfigure}
\usepackage{booktabs}
\usepackage{xcolor}
\usepackage{tikz}
\usetikzlibrary{snakes}
\usepackage{multirow}
\usepackage{epstopdf}
\usepackage{balance}
\usepackage{tablefootnote}
\usepackage{empheq} % added by christos - makes nice box for equations
\usepackage{moresize}
\usepackage{enumitem}
\setlist{nolistsep}

\usepackage[font=small,labelfont=bf,skip=0pt]{caption}

\DeclareCaptionType{copyrightbox}
\usepackage{url}

\newcounter{ALC@tempcntr}% Temporary counter for storage
\newcommand{\method}{\textsc{TenSemble2Vec}\xspace}

\newcommand{\hide}[1]{}
\newcommand{\ben}{\begin{enumerate*}}
\newcommand{\een}{\end{enumerate*}}
\newcommand{\bit}{\begin{itemize*}}
\newcommand{\eit}{\end{itemize*}}
\def\x{{\mathbf x}}

\begin{document}

\author{\IEEEauthorblockN{Jia Chen}
\IEEEauthorblockA{Dept. of Electical and Computer Engineering \\ University of Texas Rio Grande Valley\\
jia.chen@utrgv.edu}
%\email{jia.chen@utrgv.edu}}
\and
\IEEEauthorblockN{Evangelos E. Papalexakis}
\IEEEauthorblockA{Dept. of Computer Science and Engineering\\
University of California Riverside\\
epapalex@cs.ucr.edu}}
%\email{epapalex@cs.ucr.edu}}}

%\title{\method: Ensemble Node Embeddings using Tensor Decomposition}

\title{Ensemble Node Embeddings using Tensor Decomposition: A Case-Study on DeepWalk}

\maketitle

\begin{abstract}
Node embeddings have been attracting increasing attention during the past years. In this context, we propose a new ensemble node embedding approach, called  \method, by first generating multiple embeddings using the existing techniques and taking them as multiview data input of the state-of-art tensor decomposition model namely PARAFAC2 to learn the shared lower-dimensional representations of the nodes. Contrary to other embedding methods, our \method takes advantage of the 
complementary information from different methods or the same method with different hyper-parameters, which bypasses the challenge of choosing models.
%or determining the proper hyper-parameters. 
Extensive tests using real-world data validates the efficiency of the proposed method.
\end{abstract}
\section{Introduction}
\label{sec:intro}
%\reminder{Para 1node embedding techniques}
%why are they important, what are the new advances etc
Graphs are natural structure information representing the interactions between vertices/nodes, which have been broadly used in real-world scenarios \cite{goyal2018graphsurvey}. For example, in protein-protein graph, vertices indicates proteins and an edge represents a biological interconnection between a pair of proteins \cite{theocharidis2009network}; citation graph in scientific research takes individual papers as nodes and the citation relationship between two papers as an edge. Recently, learning from graphs has gained increasing attention from the research community. One of the most popular directions is node embedding, which learns latent representations of vertices for a given graph while preserving the neighborhood similarity in the original graph. Effective node embeddings empower a lot of down-streaming machine learning tasks such as node clustering, node classification, node visualization, and node recommendation, to name a few. Most node embedding techniques are based on deep learning, factorization methods, or random walks. The state-of-the-art node embedding approaches include DeepWalk \cite{perozzi2014deepwalk}, Node2Vec \cite{grover2016node2vec}, Graph Factorization \cite{ahmed2013graphfactorization}, HOPE \cite{ou2016hope}, Walklets \cite{perozzi2016walklets}, Structural Deep Network Embedding \cite{wang2016structural}, and so on.

%\reminder{Para 2 What is a problem?} Choosing the dimensionality may have an impact in the quality of embeddings. How can we do that without training data and trial-and-error?

However, finding a `good' vector representations of vertices is inherently challenging due to the difficulty of determining the dimentionality and choosing the distance metrics and properties of the graph that the learnt node vectors should preserve. For example, a proper dimension for DeepWalk ties closely to its performance. Further, which node embedding technique is a better choice remains an open question. To circumvent the challenges of the existing node embedding approaches, we propose an ensemble embedding which consolidates multiple embeddings into a single embedding. This will be realized by computing the PARAFAC2 decomposition \cite{harshman1972parafac2, perros2017spartan} of multiple datasets which are obtained from different node embeddings. The reason to choose PARAFAC2 instead of other classical multi-modal data fusion methods such as canonical polyadic (CP) decomposition, a.k.a., PARAFAC or CANDECOMP \cite{carroll1970cp}, canonical correlation analysis (CCA) \cite{hotelling1992relations}, or multiview CCA \cite{carroll1968mcca, chen2019gmcca} is fourfold: 1) CP decompostion requires all the datasets to share the number of dimension, which may not be true in many cases; 2) CCA is only capable of handling two datasets; 3) multiview CCA generalizes CCA to deal with more than two views but treats all the latent components the same; and 4) PARAFAC2 overcomes all the limitations of the aforementioned methods. 

%\reminder{Para 3 main idea} We propose an ensemble embedding which consolidates a number of embeddings (of the same or different mechanism - we can think abuot that thing in futgure work) into a single embedding that works better than or the same as its best constitutent embedding.
%\reminder{a bit on how we do it by tensor blah blah blah}
Our contributions include:
\begin{itemize}[noitemsep]
	\item {\bf Ensemble node embedding}: We develop a new ensemble node embedding scheme to overcome the shortcoming of individual embeddings.
	\item {\bf Flexibility}: Our approach has no constrains on the number of embedding datasets and the dimensions of embeddings.
	\item {\bf Experiments}: We evaluate the effectiveness of our algorithm using real-world data.
\end{itemize}
\section{Problem Formulation and Proposed Method}
Consider an undirected graph $\mathcal{G}:=\{\bf{V},\,\bf{E}\}$ consisting of $N$ nodes depicting the interactions of a network, where $\bf{V}$ collects all the nodes and  ${\bf E}\in\mathbb{R}^{N \times N}$ is the  adjacency matrix capturing the similarities between pairs of nodes satisfying $\mathbf{E}=\bf{E}^{\top}$. In this paper, our goal is to learn the node representations which preserve the network connections given by the graph $\mathcal{G}$ while transforming each node's representation from high-dimensional space $\mathbb{R}^{N}$ to a lower-dimensional space $\mathbb{R}^d$ with $d\le  N $. This will be realized by applying the existing state-of-the-art node embedding techniques to get different representations and using the PARAFAC2 \cite{harshman1972parafac2, perros2017spartan} to learn the shared representations which are our ensemble node embeddings.

\noindent{\bf Step 1: Systematic Exploration of Rich Node Embeddings.} Using solely the adjacency matrix, the first-order and second-order proximities of the node representations are commonly preserved. Using these proximity measures may not be sufficient to deliver satisfying predictive performance in some scenario. To improve the down-streaming task performance, DeepWalk implicitly preserves the higher-order proximity between the nodes by generating multiple random walks, which is implemented by maximizing the probability of observing the $2k$ nodes centered at each node in the random walk, where $k$ is the number of hops \cite{perozzi2014deepwalk}. Similarly, Node2Vec minimizes the Euclidean distance between the neighbouring node representations while preserving the higher-order proximity \cite{grover2016node2vec}. Besides, the growing research graph embedding has led to a deluge of node embedding methods including deep learning based methods \cite{wang2016structural, cao2016deepgraphreprsen}, random walk based methods \cite{perozzi2014deepwalk, grover2016node2vec}, and factorization based methods \cite{ahmed2013graphfactorization, ou2016hope, cao2015grarep}. In this paper, we will focus on DeepWalk only. The representation quality of DeepWalk is influenced by the choice of the length of node vectors which, in general, is not available. To overcome this difficulty, we will pre-define several candidates for the dimension to enable multiple node embeddings.

\noindent{\bf Step 2: Ensemble Node Representation Learning.} After conducting {\bf Step 1}, we will obtain multiple embeddings/views denoted by $\{{\bf X}_m\in{\mathbb{R}}^{N \times D_m}\}_{m=1}^M$, where $M$ is the number of embeddings from DeepWalk and $D_m$ depicts the dimension of the $m$-th embedding. Next, we will use PARAFAC2, a tensor decomposition technique, to find a shared embedding across all the $M$ embeddings. Specifically, PARAFAC2 looks for the view-specified projection matrix $\mathbf{U}_m\in\mathbb{R}^{D_m \times  R}$ and diagonal latent component importance matrix $\mathbf{S}_m\in \mathbb{R}^{R \times R}$, and shared lower-dimensional representation $\mathbf{V}\in\mathbb{R}^{N \times R}$ where $R$ is the hyperparameter specifying the number of latent components, so that $\{\mathbf{X}_m\approx \mathbf{U}_m \mathbf{S}_m \mathbf{V}^{\top}\}_{m=1}^M$. The optimization problem is as follows
\begin{align}
\label{eq:parafac2}
    \min_{\{{\bf U}_m\}, \{{\bf S}_m\}, {\bf V}}\quad &\sum_{m=1}^M \|{\bf X}_m - {\bf U}_m {\bf S}_m \bf V^\top \|_F^2\nonumber\\
    {\text{s. to}\,}\quad& {\bf U}_m = {\bf Q}_m {\bf H}, \,{\bf Q}_m^\top {\bf Q}_m={\bf I}, \,\forall{m}  
\end{align}
where ${\bf S}_m$ is diagonal, which can be solved by Alternating Least Squares approach \cite{perros2017spartan, golub2013matrix, kiers1999parafac2}.
%and admits a unique solution [].
The learnt node embedding $\bf{V}$ can be used for down-stream machine learning tasks.

\label{sec:problem}
\section{Experimental Evaluation}
\label{sec:experiments}

To validate the effectiveness of our proposed method, we will apply our approach to the well-known Karate network data \cite{zachary1977information}. Given this undirected and binary graph consisting of $34$ nodes, DeepWalk is run with the embedding dimensions $d=10, 20, 30, 40, 50, 60, 100, 200,$ and $1000$ to generate $9$ embeddings, which form the $9$ different views of the $34$ nodes and are assigned to $\{{\bf{X}}_m\}_{m=1}^{9}$ in \eqref{eq:parafac2} for \method. The clustering performance of \method on Karate network data is captured by clustering accuracy and Normalized Mutual Information (NMI) after running K-means of the obtained ensembled node embedding data, where accuracy is the number of correlately clusterd nodes divided by the total number of nodes and NMI normalizes multual information between the correct and predicted labels by the mean of the two entropy from both labels.

\begin{figure}[!htp]
	\centering 
	\includegraphics[width=0.5\textwidth]{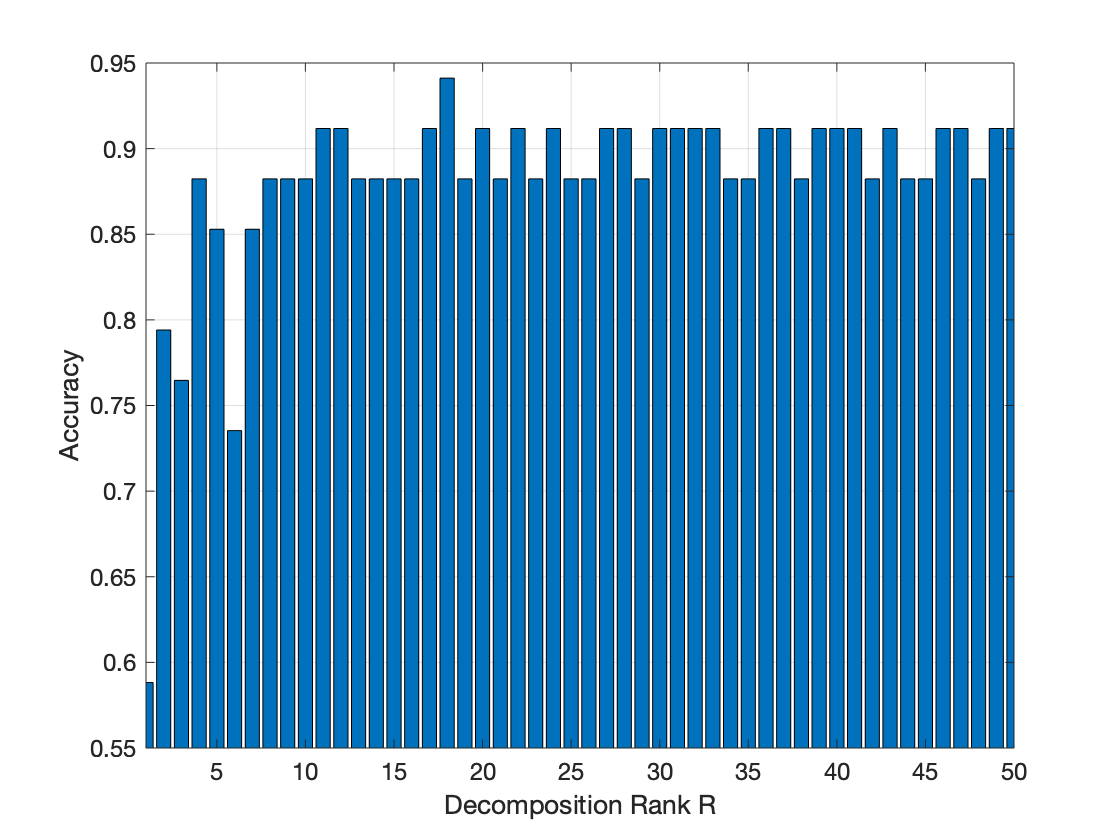} 
%	\vspace{-2pt}
	\caption{\small{Clustering accuracy of \method with different ranks.}}
	\label{fig:acc_rank}
\end{figure}

\begin{figure}[!htp]
	\centering 
	\includegraphics[width=0.5\textwidth]{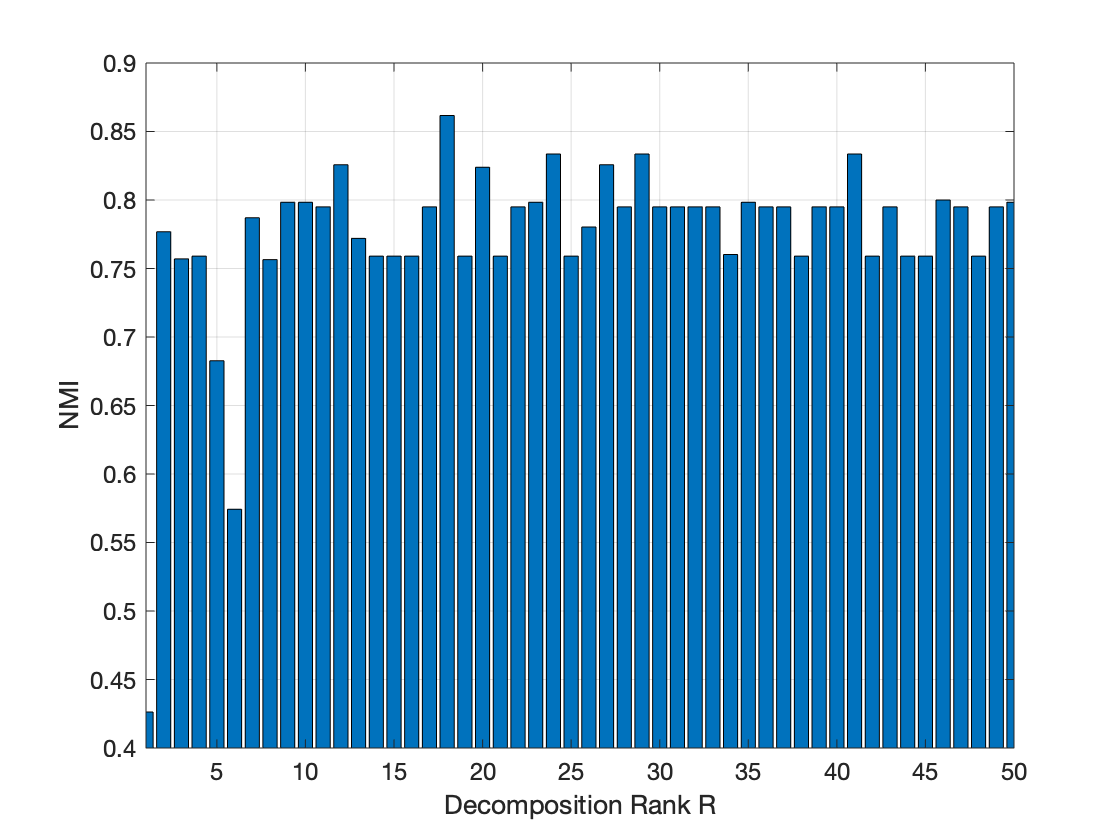} 
%	\vspace{-2pt}
	\caption{\small{Clustering NMI of \method with different ranks.}}
	\label{fig:nmi_rank}
\end{figure}

\begin{figure}[!htp]
	\centering 
	\includegraphics[width=0.5\textwidth]{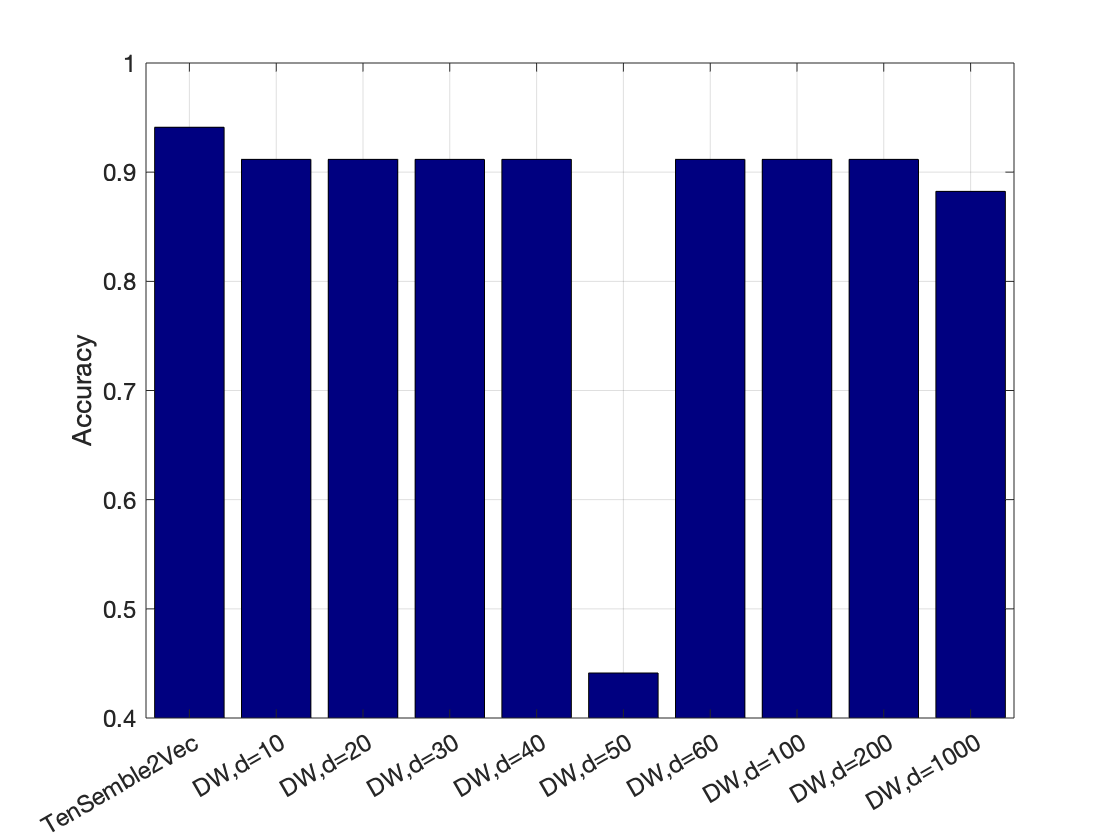} 
%	\vspace{-2pt}
	\caption{\small{Clustering accuracy comparison.}}
	\label{fig:acc_methods}
\end{figure}

\begin{figure}[!htp]
	\centering 
	\includegraphics[width=0.5\textwidth]{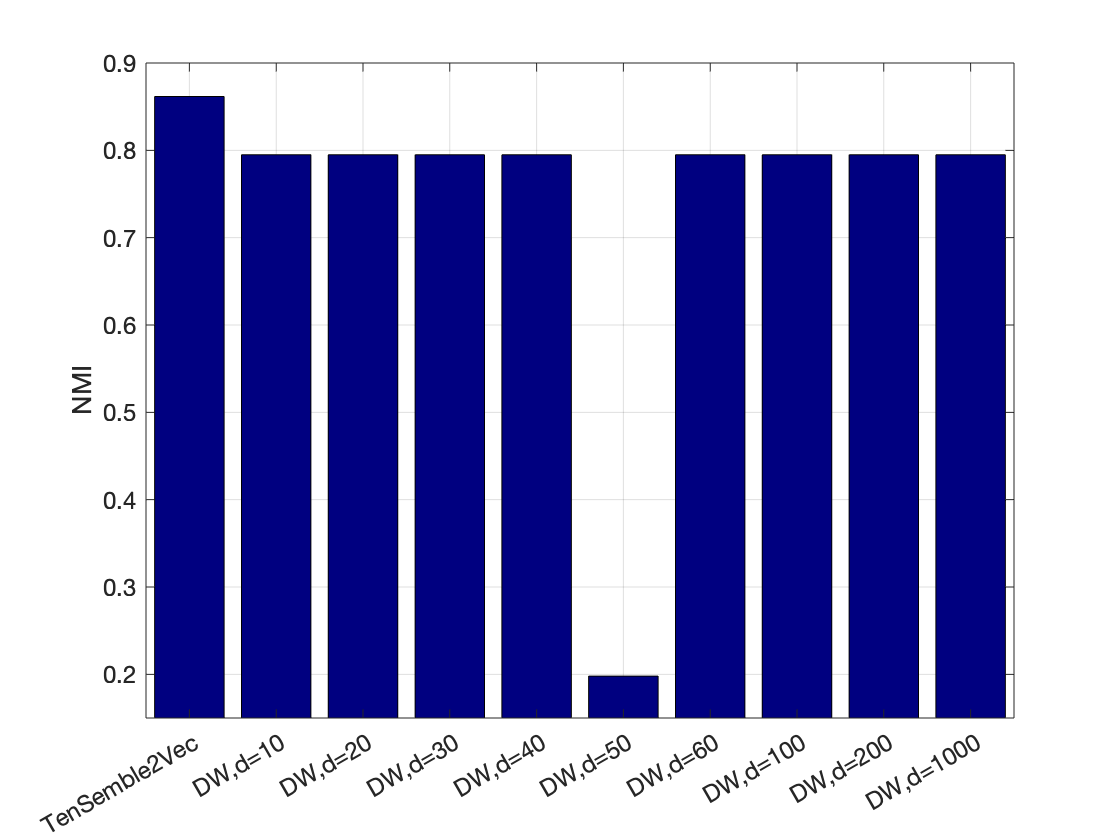} 
%	\vspace{-2pt}
	\caption{\small{Clustering NMI comparison.}}
	\label{fig:nmi_methods}
\end{figure}

First, the influence of the tensor decomposition rank $R$ to our proposed \method is investigated. Toward this end, we plot the accuracy and NMI of \method versus $R$ in Fig. \ref{fig:acc_rank} and \ref{fig:nmi_rank}, respectively, which shows that the \method achieves the best clustering performance in terms of the highest accuracy ($0.9412$) and NMI ($0.8617$) when $R=18$. Second, we compare the clustering results of \method to the DeepWalk (DW) with different embedding dimensions $d$ in Figs. \ref{fig:acc_methods} and \ref{fig:nmi_methods}. This shows that our method outperforms the existing alternatives and our ensemble node embedding works better than clustering on any single view.

\section{Conclusions}
\label{sec:conclusions}
We propose \method, a novel approach for learning latent node embeddings from an undirected graph. Using a graph adjacency matrix as input, our \method learns the node representations which preserve the structural information encoded in the adjacency by implementing different node embedding techniques to obtain different views and fusing them using PARAFAC2 to get the ensemble embedding. Promising performance on clustering Karate network data illustrates the effectiveness of our method.

Our future work will focus on using more node embedding techniques to get more views and develop an adaptive node embedding scheme to automatically decide the importance of each view. 
%Another interesting direction could be enhancing \method with self-selected tensor decomposition rank scheme.

%Ongoing work: more node embedding techniques and learn the importance of different views.

%Here we recap the list of contributions:

%Our contributions include:
%\begin{itemize}[noitemsep]
%	\item {\bf Contribution 1}: Brief description
%	\item {\bf Contribution 2}: Brief description
%	\item {\bf Contribution 3}: Brief description
%\end{itemize}

\hide{
\section{Acknowledgements}
{\scriptsize
Research was supported by the National Science Foundation Grant No. XXXXXX. Any opinions, findings, and conclusions or recommendations expressed in this material are those of the author(s) and do not necessarily reflect the views of the funding parties.
}
}
\balance
\bibliographystyle{IEEEtran}
\bibliography{BIB/vagelis_refs.bib}

\end{document}